\documentclass[11pt]{article}

\pdfoutput=1

\usepackage{acl2012}

\usepackage{times}

\usepackage{latexsym}

\usepackage{amsmath}

\usepackage{multirow}

\usepackage{url}

\usepackage{graphicx}

\usepackage{amssymb}

\usepackage{amsfonts}

\usepackage{arydshln}

\newcommand{\me}{\mathrm{e}}

\title{SimLex-999: Evaluating Semantic Models \\ With (Genuine) Similarity Estimation}

\author{Felix Hill \\
 Computer Laboratory \\
Cambridge University \\
  {\tt felix.hill@cl.cam.ac.uk} \\\And
 Roi Reichart \\
Technion, IIT\\
  {\tt roiri@ie.technion.ac.il} \\\And 
Anna Korhonen \\
 Computer Laboratory \\
Cambridge University \\
  {\tt anna.korhonen@cl.cam.ac.uk} }

\date{}

\begin{document}

\maketitle 

\begin{abstract}

We present SimLex-999, a gold standard resource for evaluating distributional semantic models that improves on existing resources in several important ways. First, in contrast to gold standards such as WordSim-353 and MEN, it explicitly quantifies \emph{similarity} rather than \emph{association} or \emph{relatedness} so that pairs of entities that are associated but not actually similar (\emph{Freud}, \emph{psychology}) have a low rating. We show that, via this focus on similarity, SimLex-999 incentivizes the development of models with a different, and arguably wider range of applications than those which reflect conceptual association. Second, SimLex-999 contains a range of concrete and abstract adjective, noun and verb pairs, together with an independent rating of concreteness and (free) association strength for each pair. This diversity enables fine-grained analyses of the performance of models on concepts of different types, and consequently greater insight into how architectures can be improved. Further, unlike existing gold standard evaluations, for which automatic approaches have reached or surpassed the inter-annotator agreement ceiling, state-of-the-art models perform well below this ceiling on SimLex-999. There is therefore plenty of scope for SimLex-999 to quantify future improvements to distributional semantic models, guiding the development of the next generation of representation-learning architectures.

\end{abstract}

\section{Introduction}

There is very little similar about coffee and cups. \emph{Coffee} refers to a plant, which is a living organism or a hot brown (liquid) drink. In contrast, a \emph{cup} is a man-made solid of broadly well-defined shape and size with a specific function relating to the consumption of liquids. Perhaps the only clear trait these concepts have in common is that they are concrete entities. Nevertheless, in what is currently the most popular evaluation gold standard for semantic similarity, WordSim(WS)-353 \cite{finkelstein2001placing}, \emph{coffee} and \emph{cup} are rated  as more `similar' than pairs such as \emph{car} and \emph{train}, which share numerous common properties (function, material, dynamic behaviour, wheels, windows etc.). Such anomalies also exist in other gold standards such as the MEN dataset \cite{bruni2012distributional}. As a consequence, these evaluations effectively penalize models for learning the evident truth that \emph{coffee} and \emph{cup} are dissimilar.

Although clearly different, \emph{coffee} and \emph{cups} are very much related. The psychological literature refers to the conceptual relationship between these concepts as \emph{association}, although it has been given a range of names including \emph{relatedness} \cite{budanitsky2006evaluating,agirre2009study}, \emph{topical similarity} \cite{hatzivassiloglou2001simfinder} and \emph{domain similarity} \cite{turney2012domain}. Association contrasts with \emph{similarity}, the relation connecting \emph{cup} and \emph{mug} \cite{tversky1977features}. At its strongest, the similarity relation is exemplified by pairs of \emph{synonyms}; words with identical referents.

Computational models that effectively capture similarity as distinct from association have numerous applications. Such models are used for the automatic generation of dictionaries, thesauri, ontologies and language correction tools \cite{cimiano2005learning,biemann2005ontology,li2006exploring}. Machine translation systems, which aim to define mappings between fragments of different languages whose meaning is similar, but not necessarily associated, are another established application \cite{he2008indirect,marton2009improved}. Moreover, since, as we establish, similarity is a cognitively complex operation that can require rich, structured conceptual knowledge to compute accurately, similarity estimation constitutes an effective proxy evaluation for general-purpose representation-learning models whose ultimate application is variable or unknown \cite{collobert:2008,baroni2010distributional}.

As we show in Section 2, the predominant gold standards for semantic evaluation in NLP do not measure the ability of models to reflect similarity. In particular, in both WS-353 and MEN, pairs of words with associated meaning, such as \emph{coffee} and \emph{cup} (rating = 6.8) \emph{telephone} and \emph{communication} (7.5) or \emph{movie} and \emph{theater} (7.7), receive a high rating regardless of whether or not their constituents are similar. Thus, the utility of such resources to the development and application of similarity models is limited, a problem exacerbated by the fact that many researchers appear unaware of what their evaluation resources actually measure.\footnote{For instance,  \newcite[pages 1,4,10]{huang2012improving} and \newcite[page 4]{reisinger2010multi} refer to MEN and/or WS-353 as `similarity datasets'. Others evaluate on both these association-based and genuine similarity-based gold standards with no reference to the fact that they measure different things \cite{medelyan2009mining,li2014obtaining}.}

While certain smaller gold standards, those of \newcite{rubenstein1965contextual} (RG) and \newcite{agirre2009study} (WS-Sim), do focus clearly on similarity, these resources suffer from other important limitations. For instance, as we show, and as is also the case for WS-353 and MEN,  state-of-the-art model performance on these evaluations has reached the average performance of a human annotator. It is common practice in NLP to define the upper limit for automated performance on an evaluation as the average human performance or inter-annotator agreement \cite{yong1999case,cunningham2005information,resnik201011}. Based on this established principle and the current evaluations, it would therefore be reasonable to conclude that the problem of representation learning, at least for similarity modelling, is approaching resolution. However, circumstantial evidence suggests that distributional models are far from perfect. For instance, we are some way from automatically-generated dictionaries, thesauri or ontologies that can be used with the same confidence as their manually-created equivalents.

Motivated by these observations, in Section 3 we present \emph{SimLex-999}, a gold standard resource for evaluating the ability of models to reflect similarity. SimLex-999 was produced by 500 paid native English speakers, recruited via Amazon Mechanical Turk\footnote{\url{www.mturk.com/}}, who were asked to rate the similarity, as opposed to association, of concepts via a simple visual interface. The choice of evaluation pairs in SimLex-999 was motivated by empirical evidence that humans represent concepts of distinct part-of-speech (POS) \cite{gentner1978relational} and conceptual concreteness \cite{hill2013quantitative} differently. While existing gold standards contain only concrete noun concepts (MEN) or cover only some of these distinctions via a random selection of items (WS-353, RG), SimLex-999 contains a principled selection of adjective, verb and noun concept pairs covering the full concreteness spectrum. This design enables more nuanced analyses of how computational models overcome the distinct challenges of representing concepts of these types.

In Section 4 we present quantitative and qualitative analyses of the SimLex-999 ratings, which indicate that participants found it unproblematic to consistently quantify the similarity of the full range of concepts and to distinguish it from association. Unlike existing datasets, SimLex-999 therefore contains a significant number of pairs, such as [\emph{movie}, \emph{theater}], which are strongly associated but receive low similarity scores.

The second main contribution of this paper, presented in Section 5, is the evaluation of state-of-the-art distributional semantic models using SimLex-999. These include the well known neuro-probabilistic language models (NLMs) of \newcite{huang2012improving}, \newcite{collobert:2008} and \newcite{mikolov2013efficient}, which we compare with traditional vector-space co-occurrence models (VSMs) \cite{turney2010frequency} and latent semantic analysis (LSA) \cite{landauer1997solution}. Our analyses demonstrate how SimLex-999 can be applied to uncover substantial differences in the ability of models to represent concepts of different types.

Despite these differences, the models we consider each share the characteristic of being better able to capture association than similarity. We show that the difficulty of estimating similarity is driven primarily by those strongly-associated pairs which have a high (association) rating in gold standards such as WS-353 and MEN, but a low similarity rating in SimLex-999. As a result of including these challenging cases, together with a wider diversity of lexical concepts in general, current models achieve notably lower scores on SimLex-999 than on existing gold standard evaluations, and well below the SimLex-999 inter-human agreement ceiling.

Finally, we explore ways in which distributional models might improve on this performance in similarity modelling. To do so, we evaluate the models on the SimLex-999 subsets of adjectives, nouns and verbs, as well as on abstract and concrete subsets and subsets of more and less strongly associated pairs (Sections 5.2.2-5.2.4). As part of these analyses, we confirm the hypothesis \cite{agirre2009study,levy2014dependency} that models learning from input informed by dependency parsing, rather than simple running-text input, yield improved similarity estimation and, specifically, clearer distinction between similarity and association. In contrast, we find no evidence for a related hypothesis \cite{agirre2009study,kiela2014systematic}, that smaller context windows improve the ability of models to capture similarity. We do, however, observe clear differences in model performance on the distinct concept types included in SimLex-999. Taken together, these experiments demonstrate the benefit of the diversity of concepts included in SimLex-999; it would not have been possible to derive similar insights by evaluating on existing gold standards.

We conclude by discussing how observations such as these can guide future research into distributional semantic models. By facilitating better-defined evaluations and finer-grained analyses, we hope that SimLex-999 will ultimately contribute to the development of models that accurately reflect human intuitions of similarity for the full range of concepts in language.

\section{Design Motivation}

In this section, we motivate the design decisions made in developing SimLex-999. We begin (2.1) by examining the distinction between similarity and association. We then show that for a meaningful treatment of similarity it is also important to take a principled approach to  both part-of-speech (POS) and conceptual concreteness (2.2). We finish by reviewing existing gold standards, and show that none enables a satisfactory evaluation of the capability of models to capture similarity (2.3).

\subsection{Similarity and Association}

The difference between association and similarity is exemplified by the concept pairs [\emph{car, bike}] and [\emph{car, petrol}]. \emph{Car} is said to be (semantically) similar to \emph{bike} and associated with (but not similar to) \emph{petrol}. Intuitively, \emph{car} and \emph{bike} can be understood as similar because of their common physical features (e.g. wheels), their common function (transport), or because they fall within a clearly definable category (modes of transport). In contrast, \emph{car} and \emph{petrol} are associated because they frequently occur together in space and language, in this case as a result of a clear functional relationship \cite{plaut1995semantic,mcrae2012semantic}.

Association and similarity are neither mutually exclusive nor independent. \emph{Bike} and \emph{car}, for instance, are related to some degree by both relations. Since it is common in both the physical world and in language for distinct entities to interact, it is relatively easy to conceive of concept pairs, such as \emph{car} and \emph{petrol}, that are strongly associated but not similar. Identifying pairs of concepts for which the converse is true is comparatively more difficult. One exception is common concepts paired with low frequency synonyms, such as \emph{camel} and \emph{dromedary}. Since the essence of association is co-occurrence (linguistic or otherwise \cite{mcrae2012semantic}), such pairs can seem, at least intuitively, to be similar but not strongly associated.

To explore the interaction between the two cognitive phenomena quantitatively, we exploited perhaps the only two existing large-scale means of quantifying similarity and association. To estimate similarity, we considered proximity in the WordNet taxonomy \cite{fellbaum1999wordnet}. Specifically, we applied the measure of \newcite{wu1994verbs} (henceforth \emph{WupSim}), which approximates similarity on a [0,1] scale reflecting the minimum distance between any two synsets of two given concepts in WordNet. WupSim has been shown to correlate well with human judgements on the similarity-focused RG dataset \cite{wu1994verbs}. To estimate association, we extracted ratings directly from the University of South Florida Free Association Database (USF) \cite{nelson2004university}. These data were generated by presenting human subjects with one of 5000 cue concepts and asking them to write the \emph{first word that comes into their head that is associated with or meaningfully related to that concept}. Each cue concept \( c \) was normed in this way by over 10 participants, resulting in a set of associates for each cue, and a total of over 72,000 \((c,a)\) pairs. Moreover, for each such pair, the proportion of participants who produced associate \(a\) when presented with cue \(c\) can be used as a proxy for the strength of association between the two concepts.

By measuring WupSim between all pairs in the USF dataset, we observed, as expected, a high correlation between similarity and association strength across all USF pairs (Spearman \( \rho= 0.65, p<0.001 \)). However, in line with the intuitive ubiquity of pairs such as \emph{car} and \emph{petrol}, of the USF pairs (all of which are associated to a greater or lesser degree) over 10\% had a WupSim score of less than 0.25. These include pairs of ontologically different entities with a clear functional relationship in the world [\emph{refrigerator, food}], which may be of differing concreteness [\emph{lung, disease}], pairs in which one concept is a small concrete part of a larger abstract category [\emph{sheriff, police}], pairs in a relationship of modification or subcategorization [\emph{gravy, boat}] and even those whose principal connection is phonetic [\emph{wiggle, giggle}]. As we show in Section 2.2, these are precisely the sort of pairs that are not contained in existing evaluation gold standards. Table 1 lists the USF noun pairs with the lowest similarity scores overall, and also those with the largest additive discrepancy between association strength and similarity.

 \begin{table}[t]\begin{center}\begin{tabular}{l|l|r|r}

Concept 1 & Concept 2 & USF & WupSim \\

\hline \emph{hatchet} & \emph{murder} & 0.013 & 0.091 \\

\emph{robbery} & \emph{jail} & 0.020 & 0.100 \\

\emph{lung} & \emph{disease} & 0.014 & 0.105 \\

\emph{burglar} & \emph{robbery}& 0.020 & 0.105\\

\hdashline \emph{sheriff} & \emph{police} & 0.333 & 0.133 \\

\emph{colonel} & \emph{army} & 0.303 & 0.111 \\

\emph{quart}& \emph{milk} & 0.462 & 0.235 \\

\emph{refrigerator} & \emph{food} & 0.424 & 0.235\\

\end{tabular}\end{center}\caption{\label{font-table} Top: Concept pairs with the lowest WupSim scores in the USF dataset overall. Bottom: Pairs with the largest discrepancy in rank between association strength (high) and WupSim (low).}\end{table}



\subsubsection{Association and similarity in NLP}

As noted in the Introduction, the similarity/association distinction is not only of interest to researchers in psychology or linguistics. Models of similarity are particularly applicable to various NLP tasks, such as lexical resource building, semantic parsing and machine translation \cite{he2008indirect,Haghighi2008Learning,marton2009improved,beltagysemantic}. Models of association, on the other hand, may be better suited to tasks such as word-sense disambiguation \cite{navigli2009word}, and applications such as text classification \cite{phan2008learning} in which the target classes correspond to topical domains such as \emph{agriculture} or \emph{sport} \cite{rose2002reuters}.

Much recent research in \emph{distributional semantics} does not distinguish between association and similarity in a principled way (see e.g. \cite{huang2012improving,reisinger2010multi,luong2013better}).\footnote{Several papers that take a knowledge-based or symbolic approach to meaning do address the similarity/association issue \cite{budanitsky2006evaluating}.} One exception is \newcite{turney2012domain}, who constructs two distributional models with different features and parameter settings, designed explicitly to capture either similarity or association. Using the output of these two models as input to a logistic regression classifier, Turney predicts whether two concepts are associated, similar or both with 61\% accuracy. However, in the absence of a gold standard covering the full range of similarity ratings (rather than a list of pairs identified as being similar or not) Turney cannot confirm directly that the similarity-focused model does indeed effectively quantify similarity.

\newcite{agirre2009study} explicitly examine the distinction between association and similarity in relation to distributional semantic models. Their study is based on the partition of WS-353 into a subset focused on similarity, which we refer to as \emph{WS-Sim}, and a subset focused on association, which we term \emph{WS-Rel}. More precisely, WS-Sim is the union of the pairs in WS-353 judged by three annotators to be similar and the set \(U\) of entirely unrelated pairs, and \emph{WS-Rel} is the union of \(U\) and pairs  judged to be associated but not similar. Agirre et al. confirm the importance of of the association/similarity distinction by showing that certain models perform relatively well on WS-Rel while others perform comparatively better on WS-Sim. However, as shown in the following section, a model need not be an exemplary model of similarity in order to perform well on WS-Sim since an important class of concept pair (associated but not similar entities) is not represented in this dataset. Therefore the insights that can be drawn from the results of the \newcite{agirre2009study} study are limited.

Several other authors touch on the similarity/association distinction in inspecting the output of distributional models \cite{andrews2009integrating,kiela2014systematic,levy2014dependency}. While the strength of the conclusions that can be drawn from such qualitative analyses is clearly limited, there appear to be two broad areas of consensus concerning similarity and distributional models:

\begin{itemize}

\item Models that learn from input annotated for syntactic or dependency relations better reflect similarity, whereas approaches that learn from running-text or bag-of-words input better model association \cite{agirre2009study,levy2014dependency}.

\item Models with larger context windows may learn representations that better capture association, whereas models with narrower windows better reflect similarity \cite{agirre2009study,kiela2014systematic}.

\end{itemize}

\subsection{Concepts, part-of-speech and concreteness}

Empirical studies have shown that the performance of both humans and distributional models depends on the POS category of the concepts learned. \newcite{gentner2006verbs} showed that children find verb concepts harder to learn than noun concepts, and \newcite{markman1997similar} present evidence that different cognitive operations are employed when comparing two nouns or two verbs. \newcite{hill2014multi} demonstrate differences in the ability of distributional models to acquire noun and verb semantics. Further, they show that these differences are greater for models that learn from both text and perceptual input (as with humans).

In addition to POS category, differences in human and computational concept learning and representation have been attributed to the effects of \emph{concreteness}, the extent to which a concept has a directly perceptible physical referent. On the cognitive side, these `concreteness effects' are well established, even if the causes are still debated \cite{paivio1991dual,hill2013quantitative}. Concreteness has also been associated with differential performance in computational text-based \cite{hill2013concreteness} and multi-modal semantic models \cite{kielaimproving}.

\subsection{Existing gold standard evaluation resources}

For brevity, we do not exhaustively review all methods that have been employed to evaluate semantic models, but instead focus on the similarity or association-based gold standards that are most commonly-applied in recent work in NLP. In each case, we consider how well the dataset satisfies one of the three following criteria:

\paragraph{Representative} The resource should cover the full range of concepts that occur in natural language. In particular, it should include cases representing the different ways in which humans represent or process concepts, and cases that are both challenging and straightforward for computational models.

\paragraph{Clearly-defined} In order for a gold standard to be diagnostic of how well a model can be applied to downstream applications, a clear understanding is needed of what exactly the gold standard measures. In particular, it must clearly distinguish between dissociable semantic relations such as association and similarity.

\paragraph{Consistent and reliable} Untrained native speakers must be able to quantify the target property consistently, without requiring lengthy or detailed instructions. This ensures that the data reflect a meaningful cognitive or semantic phenomenon, and also enables the dataset to be scaled up or transferred to other languages at minimal cost and effort.

We begin our review of existing evaluation with the gold standard most commonly-applied in current NLP research. 

\noindent 

\paragraph{\bf WordSim-353}WS-353 \cite{finkelstein2001placing} is perhaps the most commonly-used evaluation gold standard for semantic models. Despite its name, and the fact that it is often referred to as a `similarity gold standard'\footnote{See e.g. \cite{huang2012improving,bansal2014tailoring}}, in fact, the instructions given to annotators when producing WS-353 were ambiguous with respect to similarity and association. Subjects were asked to: \emph{Assign a numerical similarity score between 0 and 10 (0 = words totally unrelated, 10 = words VERY closely related) ... when estimating similarity of antonyms, consider them "similar" (i.e., belonging to the same domain or representing features of the same concept), not "dissimilar".}

\noindent 

As we confirm analytically in Section 5.2, these instructions result in pairs being rated according to association rather than similarity.\footnote{This fact is also noted by the dataset authors. See \url{www.cs.technion.ac.il/~ gabr/resources/data/wordsim353/}.} WS-353 consequently suffers two important limitations as an evaluation of similarity (which also afflict other resources to a greater or lesser degree):

\begin{enumerate}

\item Many dissimilar word pairs receive a high rating.

\item No associated but dissimilar concepts receive low ratings.

\end{enumerate}

As noted in the Introduction, an arguably more serious third limitation of WS-353 is low inter-annotor agreement, and the fact that state-of-the-art models such as those of \newcite{collobert:2008} and \newcite{huang2012improving} reach, or even surpass, the inter-annotator agreement ceiling in estimating the WS-353 scores. \newcite{huang2012improving} report a Spearman correlation of \(\rho = 0.713\) between their model output and WS-353. This is ten percentage points higher than inter-annotator agreement (\(\rho = 0.611\)) when defined as the average pairwise correlation between two annotators, as is common in NLP work \cite{pado2007flexible,reisinger2010mixture,silberer2014learning}. It could be argued that a different comparison is more appropriate: Since the model is compared to the gold-standard average across all annotators, we should compare a single annotator with the (almost) gold-standard average over all other annotators. Based on this metric the average performance of an annotator on WS-353 is  \( \rho=0.756 \), which is still only marginally better than the best automatic method.\footnote{Individual annotator responses for WS-353 were downloaded from \url{www.cs.technion.ac.il/~gabr/resources/data/wordsim353}.}

Thus, at least according to the established wisdom in NLP evaluation \cite{yong1999case,cunningham2005information,resnik201011}, the strength of the conclusions that can be inferred from improvements on WS-353 is limited. At the same time, however, state-of-the-art distributional models are clearly not perfect representation-learning or even similarity estimation engines, as evidenced by the fact they cannot yet be applied, for instance, to generate flawless lexical resources \cite{alfonseca2002extending}.

\paragraph{\bf WS-Sim} WS-Sim is the set of pairs in WS-353 identified by \newcite{agirre2009study} as either containing similar or unrelated (neither similar nor associated) concepts. The ratings in WS-Sim are mapped directly from WS-353, so that all concept pairs in WS-Sim that receive a high rating are associated and all pairs that receive a low rating are unassociated. Consequently, any model that simply reflects association would score highly on WS-Sim, irrespective of how well it captures similarity.

Such a possibility could be excluded by requiring models to perform well on WS-Sim and poorly on WS-Rel, the subset of WS-353 identified by \newcite{agirre2009study} as containing no pairs of similar concepts. However, while this would exclude models of pure association, it would not test the ability of models to effectively quantify the similarity of the pairs in WS-Sim. Put another way, the WS-Sim/WS-Rel partition could in theory resolve limitation (1) of WS-353 but it would not resolve limitation (2): Models are not tested on their ability to attribute low scores to associated but dissimilar pairs.

In fact, there are more fundamental limitations of WS-Sim as a similarity-based evaluation resource. It does not, strictly-speaking, reflect similarity at all, since the ratings of its constituent pairs were assigned by the WS-353 annotators, who were asked to estimate association, not similarity. Moreover, it inherits the limitation of low inter-annotator agreement from WS-353. The average pairwise correlation between annotators on WS-Sim is \( \rho = 0.667\), and the average correlation of a single annotator with the gold standard is only \( \rho = 0.651\), both below the performance of automatic methods \cite{agirre2009study}. Finally, the small size of WS-Sim renders it poorly representative of the full range of concepts that semantic models may be required to learn.

\paragraph{\bf Rubenstein \& Goodenough} Prior to WS-353, the smaller RG dataset, consisting of 65 pairs, was often used to evaluate semantic models. The 15 raters employed in the data collection were asked to rate the `similarity of meaning' of each concept pair. Thus RG does appear to reflect similarity rather than association. However, while limitation (1) of WS-353 is therefore avoided, RG still suffers from limitation (2): By inspection, it is clear that the low similarity pairs in RG are not associated. A further limitation is that distributional models now achieve better performance on RG (correlations of up to Person \( r = 0.86 \) \cite{hassan2011semantic}) than the reported inter-annotator agreement of \( r = 0.85 \) \cite{rubenstein1965contextual}. Finally, the size of RG renders it an even less comprehensive evaluation than WS-Sim.

\paragraph{\bf The MEN Test Collection} A larger dataset, MEN \cite{bruni2012distributional}, is used in a handful of recent studies \cite{bruni2012distributional2,bernardi2013relatedness}. As with WS-353, both of the terms \emph{similarity} and \emph{relatedness} are used by the authors when describing MEN, although the annotators were expressly asked to rate pairs according to relatedness.\footnote{http://clic.cimec.unitn.it/~elia.bruni/MEN.html}

The construction of MEN differed from RG and WS-353 in that each pair was only considered by one rater, who ranked it for relatedness relative to 50 other pairs in the dataset. An overall score out of 50 was then attributed to each pair corresponding to how many times it was ranked as more related than an alternative. However, because these rankings are based on relatedness, with respect to evaluating similarity MEN necessarily suffers from both of the limitations (1) and (2) that apply to WS-353. Further, there is a strong bias towards concrete concepts in MEN because the concepts were originally selected from those identified in an image-bank \cite{bruni2012distributional}.

\paragraph{\bf Synonym detection sets} Multiple-choice synonym detection tasks, such as the TOEFL test questions \cite{landauer1997solution}, are an alternative means of evaluating distributional models. A question in the TOEFL task consists of a cue word and four possible answer words, only one of which is a true synonym. Models are scored on the number of true synonyms identified out of 80 questions. The questions were designed by linguists to evaluate synonymy, so, unlike the evaluations considered thus far, TOEFL-style tests effectively discriminate between similarity and association. However, since they require a zero-one classification of pairs as synonymous or not, they do not test how well models discern pairs of medium or low similarity. More generally, in opposition to the  fuzzy, statistical approaches to meaning predominant in both cognitive psychology \cite{griffiths2007topics} and NLP \cite{turney2010frequency}, they do not require similarity to be measured on a continuous scale.


\section{The SimLex-999 Dataset}

Having considered the limitations of existing gold standards, in this section we describe the design of SimLex-999 in detail.

\subsection{Choice of Concepts}

\begin{figure*}[ht]  \includegraphics[width = \textwidth]{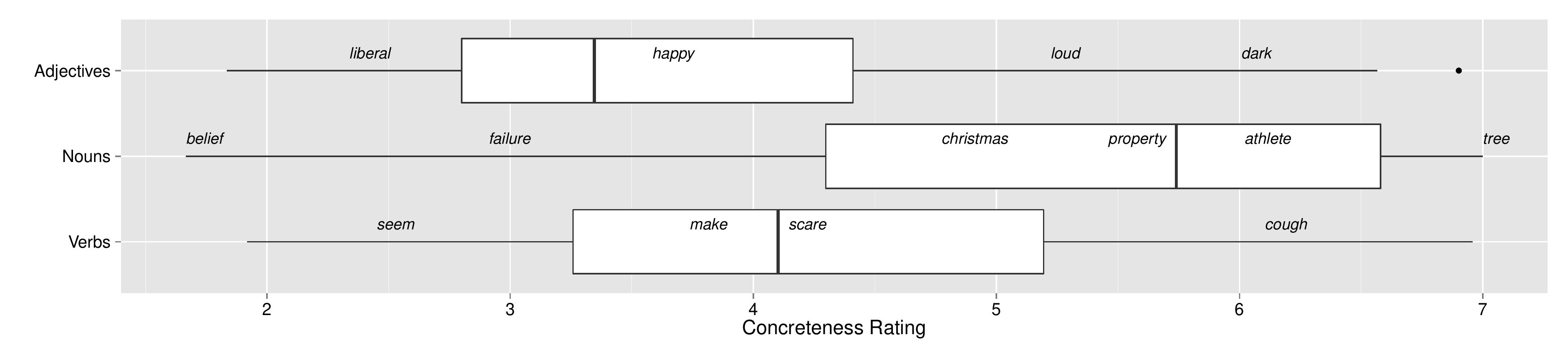}  \caption{Boxplots showing the interaction between concreteness and POS for concepts in USF. The white boxes range from the first to third quartiles and the central vertical line indicates the median.}\end{figure*}

\paragraph{Separating similarity from association}To create a test of the ability of models to capture similarity as opposed to association, we started with the \(\approx 72,000\) pairs of concepts in the USF dataset. As the output of a free-association experiment, each of these pairs is associated to a greater or lesser extent. Importantly, inspecting the pairs revealed that a good range of similarity values are represented. In particular, there were many examples of hypernym / hyponym pairs [\emph{body, abdomen}] cohyponym pairs [\emph{cat, dog}], synonyms or near synonyms [\emph{deodorant, antiperspirant}] and antonym pairs [\emph{good, evil}]. From this cohort, we excluded pairs containing a multiple-word item [\emph{hot dog, mustard}], and pairs containing a capital letter [\emph{Mexico, sun}]. We ultimately sampled 900 of the SimLex-999 pairs from the resulting cohort of pairs according to the stratification procedures outlined in the following sections.


To complement this cohort with entirely unassociated pairs, we paired up the concepts from the 900 associated pairs at random. From these random parings, we excluded those that coincidentally occurred elsewhere in USF (and therefore had a degree of association). From the remaining pairs, we accepted only those in which both concepts had been subject to the USF norming procedure, ensuring that these non-USF pairs were indeed unassociated rather than simply not normed. We sampled the remaining 99 SimLex-999 pairs from this resulting cohort of unassociated pairs.

\paragraph{POS category} In light of the conceptual differences outlined in Section 2.2, SimLex-999 includes subsets of pairs from the three principle meaning-bearing POS categories, nouns, verbs and adjectives. To classify potential pairs according to POS, we counted the frequency with which the items in each pair occurred with the three possible tags in the  POS-tagged British National Copus \cite{leech1994claws4}. To minimise POS ambiguity, which could lead to inconsistent rating, we excluded pairs containing a concept with lower than 75\% tendency towards one particular POS. This yielded three sets of potential pairs : [A,A] pairs (of two concepts whose majority tag was Adjective), [N,N] pairs and [V,V] pairs. 

\begin{figure*}[ht]  \includegraphics[width = \textwidth]{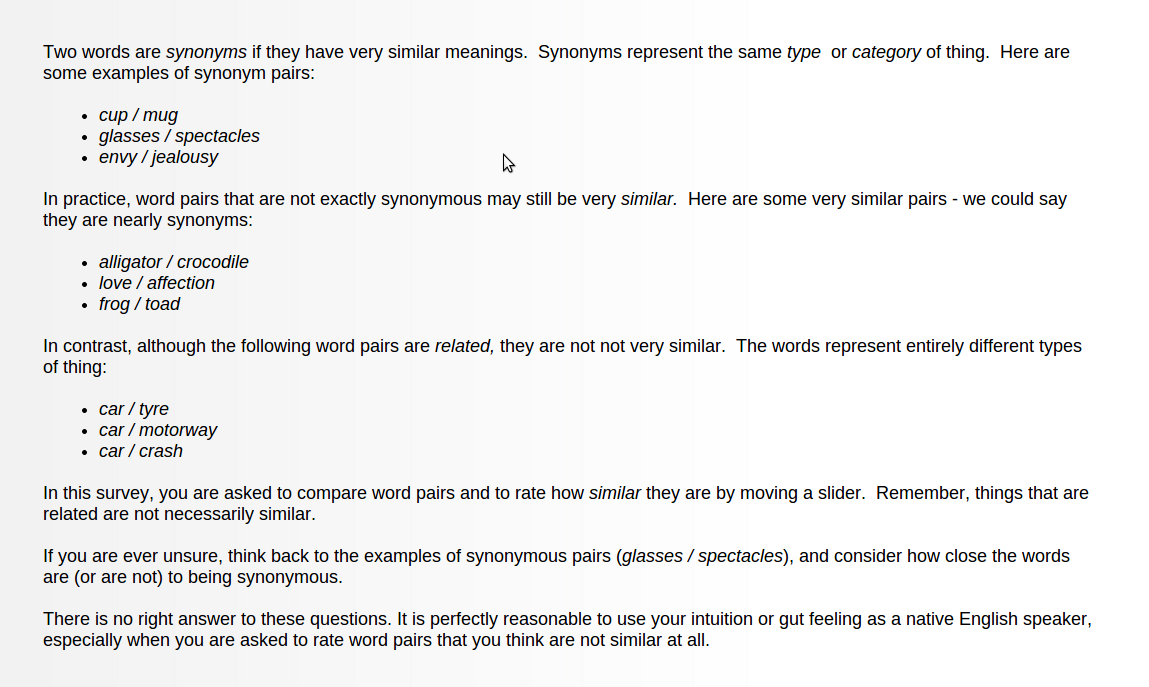}  \caption{Instructions for SimLex-999 annotators.}\end{figure*} 

Given the likelihood that different cognitive operations are employed in estimating the similarity between items of different POS-category (Section 2.2), concept pairs were presented to raters in batches defined according to POS. Unlike both WS-353 and MEN, pairs of concepts of mixed POS ([\emph{white, rabbit}], [\emph{run,marathon}]) were excluded. POS categories are generally considered to reflect very broad ontological classes \cite{fellbaum1999wordnet}. We thus felt it would be very difficult, or even counter-intuitive, for annotators to quantify the similarity of mixed POS pairs according to our instructions.

\paragraph{Concreteness} Although a clear majority of pairs in gold standards such as MEN and RG contain concrete items, perhaps surprisingly, the vast majority of adjective, noun and verb concepts in everyday language are in fact abstract \cite{hill2014multi,kielaimproving}.\footnote{According to the USF concreteness ratings, 72\% of noun or verb types in the British National Corpus are more abstract than the concept \emph{war}, a concept many would already consider quite abstract.} To facilitate the evaluation of models for both concrete and abstract concept meaning, and in light of the cognitive and computational modelling differences  between abstract and concrete concepts noted in Section 2.2, we aimed to include both concept types in SimLex-999.

Unlike the POS distinction, concreteness  is generally considered to be a gradual phenomenon. One benefit of sampling pairs for SimLex-999 from the USF dataset is that most items have been rated according to concreteness on a scale of 1-7 by at least 10 human subjects. As Figure 1 demonstrates, concreteness (as the average over these ratings) interacts with POS on these concepts: Nouns are on average more concrete than verbs which are more concrete than adjectives. However, there is also clear variation in concreteness within each POS category. We therefore aimed to select pairs for SimLex-999 that spanned the full abstract-concrete continuum within each POS category.

After excluding any pairs that contained an item with no concreteness rating, for each potential SimLex-999 pair we considered both the concreteness of the first item and the additive difference in concreteness between the two items. This enabled us to stratify our sampling equally across four classes: (\( C_1\)) Concrete first item (rating \(> 4\)) with below-median concreteness difference, (\( C_2\)) concrete first item (rating\( > 4\)) with above-median concreteness difference, (\( C_3\)) abstract first item (rating \( \leq 4\)) with below-median concreteness difference, and (\( C_4\)) abstract first item (rating \(\leq 4\)) with above-median concreteness difference.

\paragraph{Final sampling} From the associated (USF) cohort of potential pairs we selected 600 noun pairs, 200 verb pairs and 100 adjective pairs, and from the unassociated (non-USF) cohort, we sampled 66 nouns pairs, 22 verb pairs and 11 adjective pairs. In both cases, the sampling was stratified such that, in each POS subset, each of the four concreteness classes \(C_1 - C_4\) was equally represented.

\subsection{Question Design}

The annotator instructions for SimLex-999 are shown in Figure 2. We did not attempt to formalise the notion of similarity, but rather introduce it via the well-understood idea of synonymy, and in contrast to association. Even if a formal characterisation of similarity existed, the evidence in Section 2 suggests that the instructions would need separate cases to cover different concept types, increasing the difficulty of the rating task. Therefore we preferred to appeal to intuition on similarity, and to verify post-hoc that subjects were able to interpret and apply the informal characterization consistently for each concept type.

Immediately following the instructions in Figure 2, participants were presented with two 'checkpoint' questions, one with abstract examples and one with concrete examples. In each case the participant was required to identify the \emph{most similar} pair from a set of three options, all of which were associated, but only one of which was clearly similar (e.g. [\emph{bread, butter}] [\emph{bread, toast}] [\emph{stale, bread}]). After this, the participants began rating pairs in groups of 6 or 7 pairs by moving a slider, as shown in Figure 3.

This group size was chosen because the (relative) rating a set of pairs implicitly requires pairwise comparisons between all pairs in that set. Therefore, larger groups would have significantly increased the cognitive load on the annotators. Another advantage of grouping was the clear break (submitting a set of ratings and moving to the next page) between the tasks of rating adjective, noun and verb pairs. For better inter-group calibration, from the second group onwards the last pair of the previous group became the first pair of the present group, and participants were asked to re-assign the rating previously attributed to the first pair before rating the remaining new items.

\begin{figure}

    \includegraphics[width=0.5\textwidth]{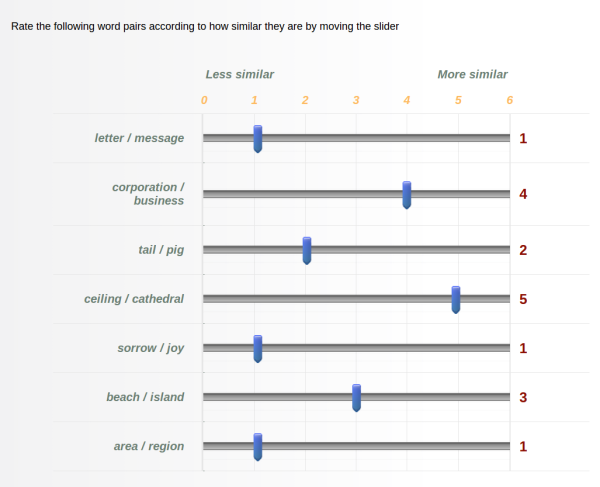}

  \caption{A group of noun pairs to be rated by moving the sliders. The rating slider was initially at position 0, and it was possible to attribute a rating of 0, although it was necessary to have actively moved the slider to that position to proceed to the next page.}

\end{figure}


\subsection{Context-free rating}

As with MEN, WS-353 and RG, SimLex-999 consists of pairs of concept words together with a numerical rating. Thus, unlike in the small evaluation constructed by \newcite{huang2012improving}, words are not rated in a phrasal or sentential context. Such meaning-in-context evaluations are motivated by a desire to disambiguate words that otherwise might be considered to have multiple senses.

We did not attempt to construct an evaluation based on meaning-in-context for several reasons. First, determining the set of senses for a given word, and then the set of contexts that represent those senses, introduces a high degree of subjectivity into the design process. Second, ensuring that a model has learned a high quality representation of a given concept would have required evaluating that concept in each of its given contexts, necessitating many more cases and a far greater annotation effort. Third, in the (infrequent) case that some concept \(c_1\) in an evaluation pair \((c_1,c_2)\) is genuinely (etymologically) polysemous, \( c_2 \) can provide sufficient context to disambiguate \(c_1\).\footnote{This is supported by the fact that the WordNet-based methods that perform best at modeling human ratings  model the similarity between concepts \( c_1 \) and \( c_2 \) as the minimum of all pairwise distances between the senses of \(c_1\) and the senses of \(c_2\) \cite{resnik1995using,pedersen2004wordnet}.} Finally, the POS grouping of pairs in the survey can also serve to disambiguate in the case that the conflicting senses of the polysemous concept are of differing POS category.

\subsection{Questionnaire structure}

Each participant was asked to rate 20 groups of pairs on a 0-6 scale of integers (non-integral ratings were not possible). Checkpoint multiple-choice questions were inserted at points between the 20 groups in order to ensure the participant had retained the correct notion of similarity. In addition to the checkpoint of three noun pairs presented before the first group (which contained noun pairs), checkpoint questions containing adjective pairs were inserted before the first adjective group and checkpoints of three verb pairs were inserted before the first verb group.

From the 999 evaluation pairs, 14 noun pairs, 4 verb pairs and 2 adjective pairs were selected as a \emph{consistency set}. The dataset of pairs was then partitioned into 10 tranches, each consisting of 119 pairs, of which 20 were from the consistency set and the remaining 99 unique to that tranche. To reduce workload, each annotator was asked to rate the pairs in a single tranche only. The tranche itself was divided into 20 groups, with each group consisting of 7 pairs (with the exception of the last group of the 20, which had 6). Of these 7 pairs, the first pair was the last pair from the previous group, and the second pair was taken from the consistency set. The remaining pairs were unique to that particular group and tranche. The design enabled control for possible systematic differences between annotators and tranches, which could be detected by variation on the consistency set.

\subsection{Participants}

500 residents of the USA were recruited from Mechanical Turk, each with at least 95\% approval rate for work on the web service. Each participant was required to check a box confirming that he or she was a native speaker of English and warned that work would be rejected if the pattern of responses indicated otherwise. The participants were distributed evenly to rate pairs in one of the ten question tranches, so that each pair was rated by approximately 50 subjects. Participants took between 8 and 21 minutes to rate the 119 pairs across the 20 groups, together with the checkpoint questions.

\subsection{Post-processing}

In order to correct for systematic differences in the overall calibration of the rating scale between respondents, we measured the average (mean) response of each rater on the consistency set. For 32 respondents, the absolute difference between this average and the mean of all such averages was greater than one (though never greater than two); i.e. 32 respondents demonstrated a clear tendency to rate pairs as either more or less similar than the overall rater population. To correct for this bias, we increased (or decreased) the rating of such respondents for each pair by one, except in cases where they had given the maximum rating, six (or minimum rating, zero). This adjustment, which ensured that the average response of each participant was within one of the mean of all respondents on the consistency set, resulted in a small increase to the inter-rater agreement on the dataset as a whole.

After controlling for systematic calibration differences, we imposed three conditions for the responses of a rater to be included in the final data collation.  First, the average pairwise Spearman correlation of responses with all other responses for a participant could not be more than one standard deviation below the mean of all such averages. Second, the increase in inter-rater agreement when a rater was excluded from the analysis needed to be smaller than at least 50 other raters (i.e. 10\% of raters were excluded on this criterion). Finally, we excluded the 6 participants who got one or more of the checkpoint questions wrong. A total of 99 participants were excluded based on one or more of these conditions, but no more than 16 from any one tranche (so that each pair in the final dataset was rated by a minimum of 36 raters).

\section{Analysis of Dataset}

In this section we analyse the responses of the SimLex-999 annotators and the resulting ratings. First, by considering inter-annotator agreement we examine the consistency with which annotators were able to apply the characterization of similarity outlined in the instructions to the range of concept types in SimLex-999. Second, we verify that a valid notion of similarity was understood by the annotators, in that they were able to accurately separate similarity from association.

\subsection{Inter-annotator agreement}

\begin{figure*}[ht]  \includegraphics[width = \textwidth, height = 3.5cm]{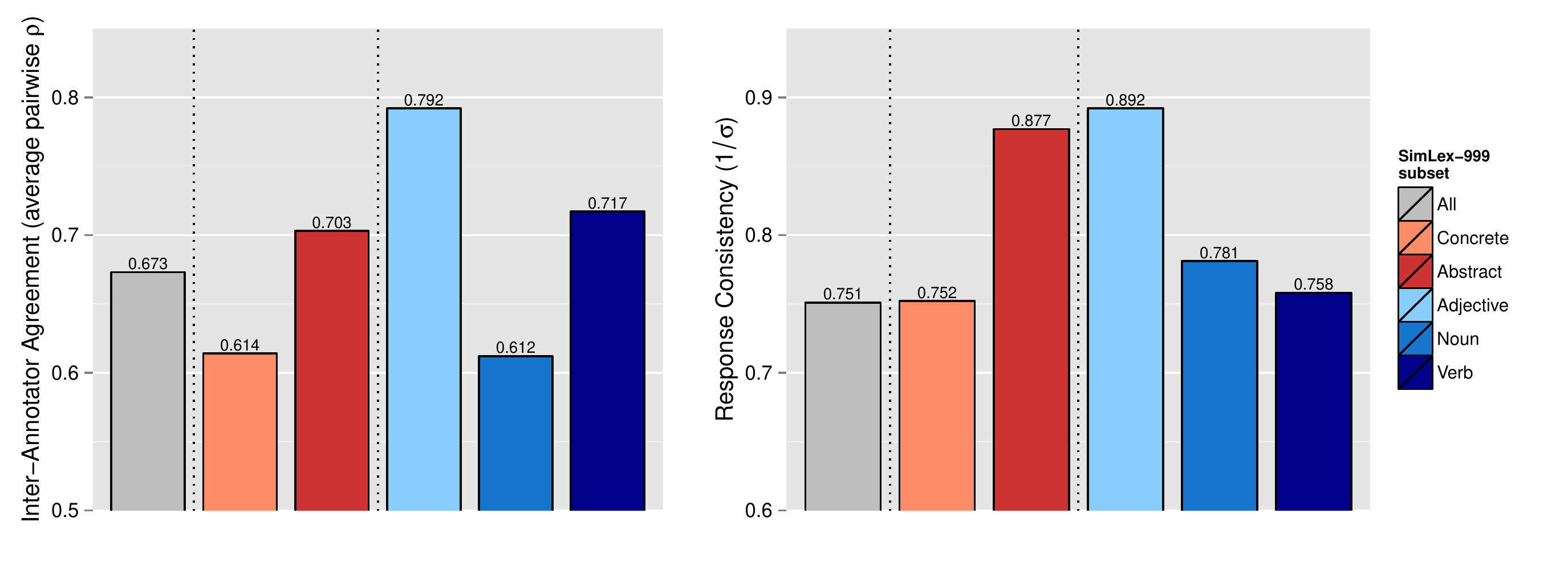}  \caption{{\bf Left:} Inter-annotator agreement, measured by average pairwise Spearman \(\rho\) correlation, for ratings of concepts of different types in SimLex-999. {\bf Right:} Response consistency, reflecting the standard deviation of annotator ratings for each pair, averaged over all pairs in the concept category.}\end{figure*}

As in previous annotation or data collection for computational semantics \cite{pado2007flexible,reisinger2010mixture,silberer2014learning} we computed the inter-rater agreement as the average of pairwise Spearman \(\rho\) correlations between the ratings of all respondents. Overall agreement was \(\rho=0.67\). This compares favourably with the agreement on WS-353 (\(\rho=0.61\) using the same method). The design of the MEN rating system precludes a conventional calculation of inter-rater agreement \cite{bruni2012distributional2}. However, two of the creators of MEN who independently rated the dataset achieved an agreement of \(\rho=0.68\).\footnote{Reported at http://clic.cimec.unitn.it/~elia.bruni/MEN. It is reasonable to assume that actual agreement on MEN may be somewhat lower than 0.68 given the small sample size and the expertise of the raters.}

The SimLex-999 inter-rater agreement suggests that participants were able to understand the (single) characterization of similarity presented in the instructions and to apply it to concepts of various types consistently. This conclusion was supported by inspection of the brief feedback offered by the majority of annotators in a final text field in the questionnaire: 78\% expressed sentiment that the test was clear, easy to complete or some similar sentiment.

Interestingly, as shown in Figure 4 (left), agreement was not uniform across the concept types. Contrary to what might be expected given established concreteness effects \cite{paivio1991dual}, we observed not only higher inter-rater agreement but also less per-pair variability for abstract rather than concrete concepts\footnote{Per-pair variability was measured by calculating the standard deviation of responses for each pair, and averaging these scores across the pairs of a each concept type.}.

Strikingly, the highest inter-rater consistency and lowest per-pair variation (defined as the inverse of the standard deviation of all ratings for that pair) was observed on adjective pairs. While we are unsure exactly what drives this effect, a possible cause is that many pairs of adjectives in SimLex-999 cohabit a single salient, one-dimensional scale (\emph{freezing > cold > warm > hot}). This may be a consequence of the fact that many pairs in SimLex-999 were selected (from USF) to have a degree of association. On inspection, pairs of nouns and verbs in SimLex-999 do not appear to occupy scales in the same way, possibly since concepts of these POS categories come to be associated via a more diverse range of relations. It seems plausible that humans are able to estimate the similarity of scale-based concepts more consistently than pairs of concepts related in a less uni-dimensional fashion.

Regardless of cause, however, the high agreement on adjectives is a satisfactory property of SimLex-999. Adjectives exhibit various aspects of lexical semantics that have proved challenging for computational models, including antonymy, polarity \cite{williams2009predicting} and sentiment \cite{wiebe2000learning}. To approach the high level of human confidence on the adjective pairs in SimLex-999, it may be necessary to focus particularly on developing automatic ways toi capture these phenomena.

\subsection{Response validity: Similarity not association}

Inspection of the SimLex-999 ratings indicated that pairs were indeed evaluated according to similarity rather than association. Table 2 includes examples that demonstrate a clear dissociation between the two semantic relations.

 \begin{table*}[ht]\begin{center}\begin{tabular}{l|l|c|r|r|r|r}

C1 & C2 & POS & USF* & USF rank (of 999) & SimLex & SimLex rank (of 999) \\

\hline \emph{dirty} & \emph{narrow} & A & 0.00 & 999 & 0.30 & 996 \\

\emph{student} & \emph{pupil} & N & 6.80 & 12 & 9.40 & 12 \\

\emph{win} & \emph{dominate} & V & 0.41 & 364 & 5.68 & 361 \\

\hdashline \emph{smart} & \emph{dumb} & A & 2.10 & 92 & 0.60 & 947 \\

\emph{attention} & \emph{awareness} & N &  0.10 & 895 & 8.73 & 58 \\

\emph{leave} & \emph{enter} & V & 2.16 & 89 & 1.38 & 841 \\

\end{tabular}\end{center}\caption{\label{font-table} {\bf Top:  Similarity aligns with association} Pairs with a small difference in rank between USF (association) and SimLex-999 (similarity) scores for each POS category. {\bf Bottom: Similarity contrasts with association} Pairs with a high difference in rank for each POS category. *Note that the distribution of USF association scores on the interval [0,10] is highly skewed towards the lower bound in both SimLex-999 and the USF dataset as a whole.}\end{table*}

\begin{figure*}[ht]  \includegraphics[width =\textwidth, height=4cm]{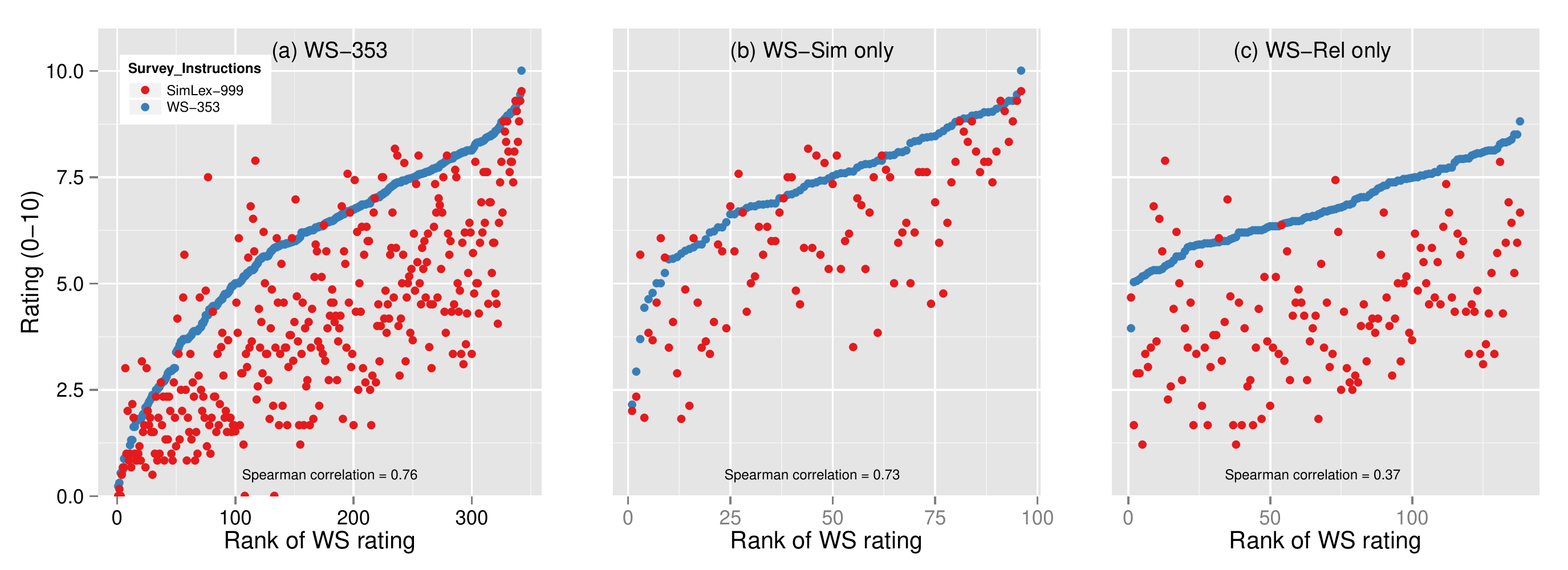}  \caption{{\bf(a)} Pairs rated by WS-353 annotators (blue points, ranked by rating) and the corresponding rating of annotators following the SimLex-999 instructions (red points). {\bf(b-c)} The same analysis, restricted to pairs in the WS-Sim or WS-Rel subsets of WS-353.}\end{figure*}

To verify this effect quantitatively, we recruited 100 additional participants to rate the WS-353 pairs, but following the SimLex-999 instructions and question format. As shown in Fig 5(a), there were clear differences between these new ratings and the original WS-353 ratings. In particular, a high proportion of pairs was given a lower rating by subjects following the SimLex-999 instructions than those following the WS-353 guidelines: The mean SimLex rating was 4.07 compared with 5.91 for WS-353.

This was consistent with our expectations that pairs of associated but dissimilar concepts would receive lower ratings based on the SimLex-999 than on the WS-353 instructions while pairs that were both associated and similar would receive similar ratings in both cases. To confirm this, we compared the WS-353 and SimLex-999-based ratings on the subsets WS-Rel and WS-Sim, which were hand-sorted by \newcite{agirre2009study} to include pairs connected by association (and not similarity) and those connected by similarity (but possibly also association) respectively.

As shown in Figure 5(b-c), the correlation between the SimLex-999-based and WS-353 ratings was notably higher (\(\rho=0.73\)) on the WS-Sim subset than the WS-Rel subset (\(\rho=0.38\)). Specifically, the tendency of subjects following the SimLex-999 instructions to assign lower ratings than those following the WS-353 instructions was far more pronounced for pairs in WS-Sim (Figure 5(b)) than for those in WS-Rel (5(c)). This observation suggest that the associated but dissimilar pairs in WS-353 were an important driver of the overall lower mean for SimLex-999-based ratings, and thus provide strong evidence that the SimLex-999 instructions do indeed enable subjects to effectively distinguish similarity from association.


\section{Evaluating Models with SimLex-999}

In this section, we demonstrate the applicability of SimLex-999 by analysing the performance of various distributional semantic models in estimating the new ratings. The models were selected to cover the main classes of representation learning architectures \cite{baroni2014don}: Vector space co-occurrence (counting) models and neural language models (NLM)s \cite{bengio2003}. We first show that SimLex-999 is notably more difficult for state-of-the-art models to estimate than existing gold standards. We then conduct more focused analyses on the various concept subsets defined in SimLex-999, exploring possible causes for the comparatively low performance of current models and, in turn, demonstrating how SimLex-999 can be applied to investigate such questions.

\subsection{Semantic models}

\paragraph{\bf Collobert \& Weston}

\newcite{collobert:2008} apply the architecture of an NLM to learn a word representations \(v_w\) for each word \(w\) in some corpus vocabulary \( V\). Each sentence \( s\) in the input text is represented by a matrix containing the vector representations of the words in \(s\) in order. The model then computes output scores \(f(s) \) and \(f(s^w) \), where \(s^w\) denotes an `incorrect' sentence created from \(s\) by replacing its last word with some other word \( w\) from \(V\). Training involves updating the parameters of the function \(f\) and the entries of the vector representations \(v_w\) such that  \(f(s)\) is larger than \(f(s^w) \) for any \(w\) in \(V\), other than the correct final word of \(s\). This corresponds to minimising the sum of the following sentence objectives \( C_s\) over all sentences in the input corpus, which is achieved via (mini-batch) stochastic gradient descent:

\[ C_{s}  = \sum_{w \in V} max(0,1-f(s) + f(s^w)). \]

The relatively low-dimension, dense (vector) representations learned by this model and the other NLMs introduced in this section are sometimes referred to as \emph{embeddings} \cite{turian2010word}. \newcite{collobert:2008} train their models on 852 million words of text from a 2007 dump of Wikipedia and the RCV1 Corpus \cite{lewis2004rcv1} and use their embeddings to achieve state-of-the-art results on a variety of NLP tasks. We downloaded the embeddings directly from the authors' webpage.\footnote{http://ml.nec-labs.com/senna/}

 \paragraph{\bf Huang et al.}

\newcite{huang2012improving} present a NLM that learns word embeddings to maximise the likelihood of predicting the last word in a sentence \(s\) based on (i) the previous words in that sentence (local context - as with \newcite{collobert:2008}) and (ii) the document \( d\) in which that word occurs (global context). As with \newcite{collobert:2008}, the model represents input sentences as a matrix of word embeddings. In addition, it represents documents in the input corpus as single-vector averages over all word embeddings in that document. It can then compute scores \(g(s,d )\) and \(g(s^w, d) \), where as before \(s^w\) is a sentence with an `incorrect' randomly-selected last word. Training is again by stochastic gradient descent, and corresponds to minimising the sum of the sentence objectives \(C_{s,d} \) over all of the sentences in the corpus:

\[ C_{s,d}  = \sum_{w \in V} max(0,1-g(s,d) + g(s^w,d)). \]

The combination of local and global contexts in the objective encourages the final word embeddings to reflect aspects of both the meaning of nearby words and of the documents in which those words appear. When learning from 990m words of wikipedia text, Huang et al. report a Spearman correlation of \(\rho = 71.3\) between the cosine similarity of their model embeddings and the WS-353 scores, which constitutes state-of-the-art performance for a NLM model on that dataset. We downloaded these embeddings from the authors' webpage.\footnote{\url{www.socher.org}.}

\paragraph{\bf Mikolov et al.}

\newcite{mikolov2013efficient} present an architecture that learns word embeddings similar to those of standard NLMs but with no non-linear hidden layer (resulting in a simpler scoring function). This enables faster representation learning for large vocabularies. Despite this simplification, the embeddings achieve state-of-the-art performance on several semantic tasks including sentence completion and analogy modelling \cite{mikolov2013efficient,mikolov2013distributed}.

For each word type \(w\) in the vocabulary \(V\), the model learns both a `target-embedding' \( r_{w} \in \mathbb{R}^d\) and a `context-embedding' \(\hat{r}_{w} \in \mathbb{R}^d\) such that, given a target word, its ability to predict nearby context words is maximized. The probability of seeing context word \(c\) given target \(w\) is defined as:

\[p(c|w)  = \frac{\me^{\hat{r}_{c} \cdot r_{w}}}{\sum_{v \in V} \me^{\hat{r}_v\cdot r_{w}}}    \]

The model learns from a set of (target-word, context-word) pairs, extracted from a corpus of sentences as follows. In a given sentence \(s\) (of length \(N\)), for each position \( n \leq N\), each word \(w_n\) is treated in turn as a target word. An integer \( {t(n)} \) is then sampled from a uniform distribution on \( \{1, \dots k \} \), where \(k > 0\) is a predefined maximum context-window parameter. The pair tokens \( \{(w_n, w_{n+j}): -{t(n)}\leq j \leq {t(n)}, w_i \in s \}\) are then appended to the training data. Thus, target/context training pairs are such that (i) only words within a \(k\)-window of the target are selected as context words for that target, and (ii) words closer to the target are more likely to be selected than those further away.

The training objective is then to maximize the log probability \( T\), defined below, across of all such examples from \(s\), and then across all sentences in the corpus. This is achieved by stochastic gradient descent.

\[ T = \frac{1}{N} \sum_{n=1}^{N} \sum_{-{t(n)}\leq j \leq {t(n)}, j\neq 0} log(  p(w_{n+j}|w_{n}) ) \]

As with other NLMs, Mikolov et al.'s model captures conceptual semantics by exploting the fact that words appearing in similar linguistic contexts are likely to have similar meanings. Informally, the model adjusts its embeddings to increase the probability of observing the training corpus. Since this probability increases with \(p(c|w)\), and \(p(c|w)\) increases with the dot product \( \hat{r}_c\cdot r_{w} \), the updates have the effect of moving each target-embedding incrementally `closer' to the context-embeddings of its collocates. In the target-embedding space, this results in embeddings of concept words that regularly occur in similar contexts moving closer together.

We use the author's Word2vec software in order to train their model and use the target embeddings in our evaluations. We experimented with embeddings of dimension 100, 200, 300, 400 and 500 and found that 200 gave the best performance on both WS-353 and SimLex-999.

\paragraph{\bf Vector Space Model (VSM)}

As an alternative to NLMs, we constructed a vector space model following the guidelines for optimal performance outlined by \newcite{kiela2014systematic}. After extracting the 2000 most frequent word tokens in the corpus that are not in a common list of stopwords\footnote{Taken from the Python Natural Language Toolkit \cite{bird2006nltk}.} as features, we populated a matrix of co-occurrence counts with a row for each of the concepts in some pair in our evaluation sets, and a column for each of the features. Co-occurrence was counted within a specified window size, although never across a sentence boundary. This resulting matrix was then weighted according to Pointwise Mutual Information (PMI) \cite{recchia2009more}. The rows of the resulting matrix constitute the vector representations of the concepts.

\paragraph{\bf LSA} Our LSA model was constructed as per the VSM, but with an extra dimensionality-reduction step. As described by \newcite{landauer1997solution}, we applied Singular Value Decomposition (SVD) \cite{golub1970singular} to the PMI-weighted VSM matrix, reducing the dimension of each concept representation to 300 (which yielded best results after experimenting, as before, with 100-500 dimension vectors).

\vspace{1\baselineskip}

\noindent 
For each model described in this section, we calculate similarity as the cosine similarity between the (vector) representations learned by that model.

\subsection{Results}

In experimenting with different models on SimLex-999, we aimed to answer the following questions: (i) How well do the established models perform on SimLex-999 versus on existing gold standards? (ii) Are any observed differences caused by the potential of different models to measure similarity vs. association? (iii) Are there interesting differences in ability of models to capture similarity between adjectives vs nouns vs verbs? (iv) In this case, are the observed differences driven by concreteness, and its interaction with POS, or are other factors also relevant?

\paragraph{\bf Overall performance on SimLex-999}

\begin{figure*}[ht]  \includegraphics[width = \textwidth, height=4cm]{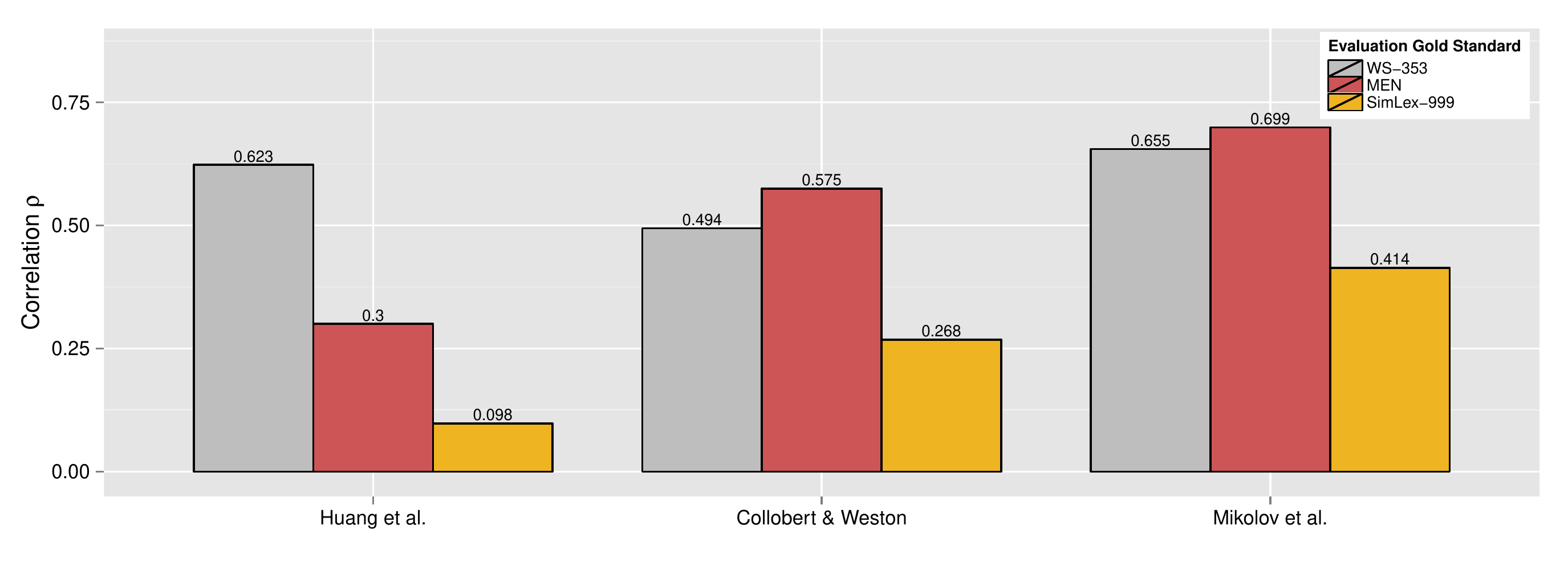} \caption{Performance of NLMs on WS-353, MEN and SimLex-999. All models are trained on Wikipedia; note that as Wikipedia is constantly growing, the \protect\newcite{mikolov2013efficient} model exploited slightly more training data ($\approx$1000m tokens) than the \protect\newcite{huang2012improving} model ($\approx$990m), which in turn exploited more than the \protect\newcite{collobert:2008} model ($\approx$852m).}\end{figure*}

\begin{figure*}[ht]  \includegraphics[width = \textwidth,height=4cm]{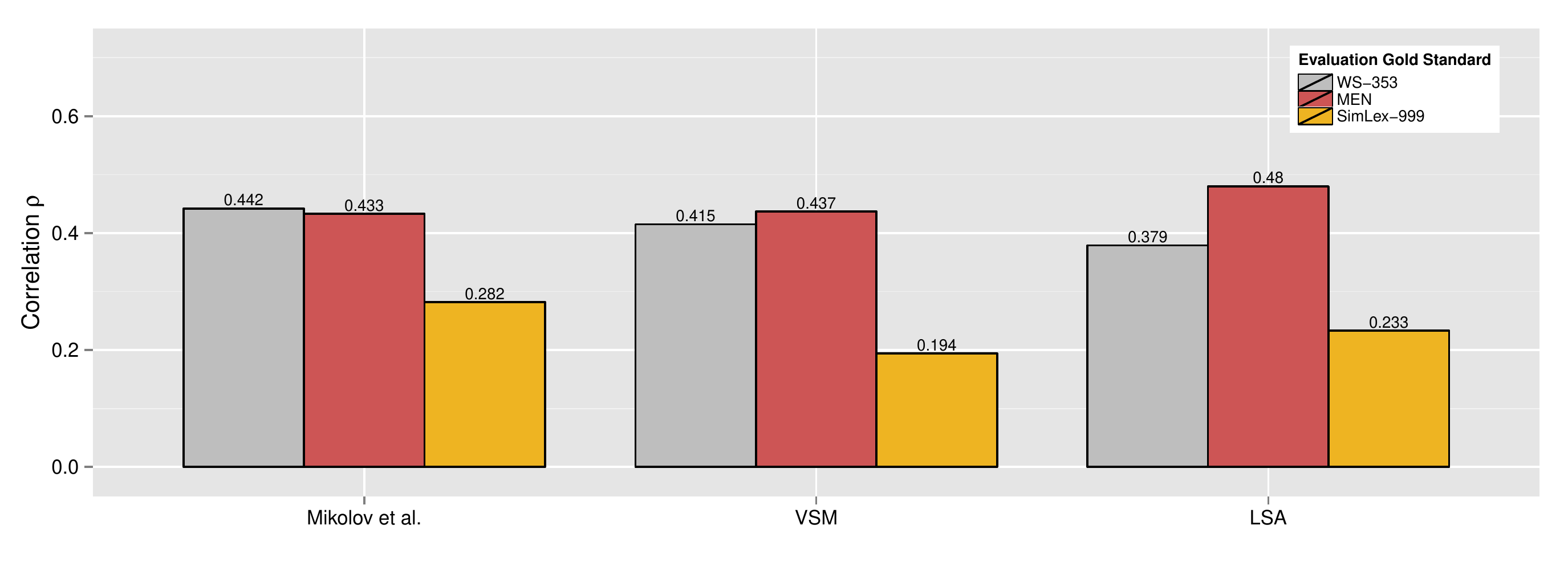}  \caption{Comparison between the leading NLM, \emph{Mikolov et al.}, the vector space model, \emph{VSM}, and the \emph{LSA} model. All models were trained on the $\approx$150m word RCV1 Corpus \protect\cite{lewis2004rcv1}.}\end{figure*}

Figure 6 shows the performance of the NLMs on SimLex-999 versus on comparable datasets, measured by Spearman's \(\rho\) correlation. All models estimate the ratings of MEN and WS-353 more accurately than SimLex-999. The \newcite{huang2012improving} model performs well on WS-353\footnote{This score, based on embeddings downloaded from the authors' webpage, is notably lower than the score reported in \cite{huang2012improving} mentioned in Section 5.1.}, but is not very robust to changes in evaluation gold standard, and performs worst of all the models on SimLex-999. Given the focus of the WS-353 ratings, it is tempting to explain this by concluding that the global context objective leads the \newcite{huang2012improving} model to focus on association rather than similarity. However, the true explanation may be less simple, since the \newcite{huang2012improving} model performs weakly on the association-based MEN dataset. The \newcite{collobert:2008} model is more robust across WS-353 and MEN, but still does not match the performance of the \newcite{mikolov2013efficient} model on SimLex-999.

Figure 7 compares the best performing NLM model \cite{mikolov2013efficient} with the VSM and LSA models.\footnote{We conduct this comparison on the smaller RCV1 Corpus \cite{lewis2004rcv1} because training the VSM and LSA models is comparatively slow.}    In contrast to recent results that emphasize the superiority of NLMs over alternatives \cite{baroni2014don}, we observed no clear advantage for the NLM over the VSM or LSA when considering the association-based gold standards WS-353 and MEN together. While the NLM is the strongest performer on WS-353, LSA is the strongest performer on MEN. However, the NLM model performs notably better than the alternatives at modelling similarity, as measured by SimLex-999.

Comparing all models in Figures 6 and 7 suggests that SimLex-999 is notably more challenging to model than the alternative datasets, with correlation scores ranging from 0.098 to 0.414. Thus, even when state-of-the-art models are trained for several days on the largest input corpora we are aware of (Figure 6)\footnote{Training times reported by \newcite{huang2012improving}, and for \newcite{collobert:2008} at \url{http://ronan.collobert.com/senna/}.}, their performance on SimLex-999 is well below the inter-annotator agreement of 0.67. This suggests that there is ample scope for SimLex-999 to guide the development of improved models.

\paragraph{\bf Modeling similarity vs. association}

\begin{figure*}[ht]  \includegraphics[width = \textwidth,height=4cm]{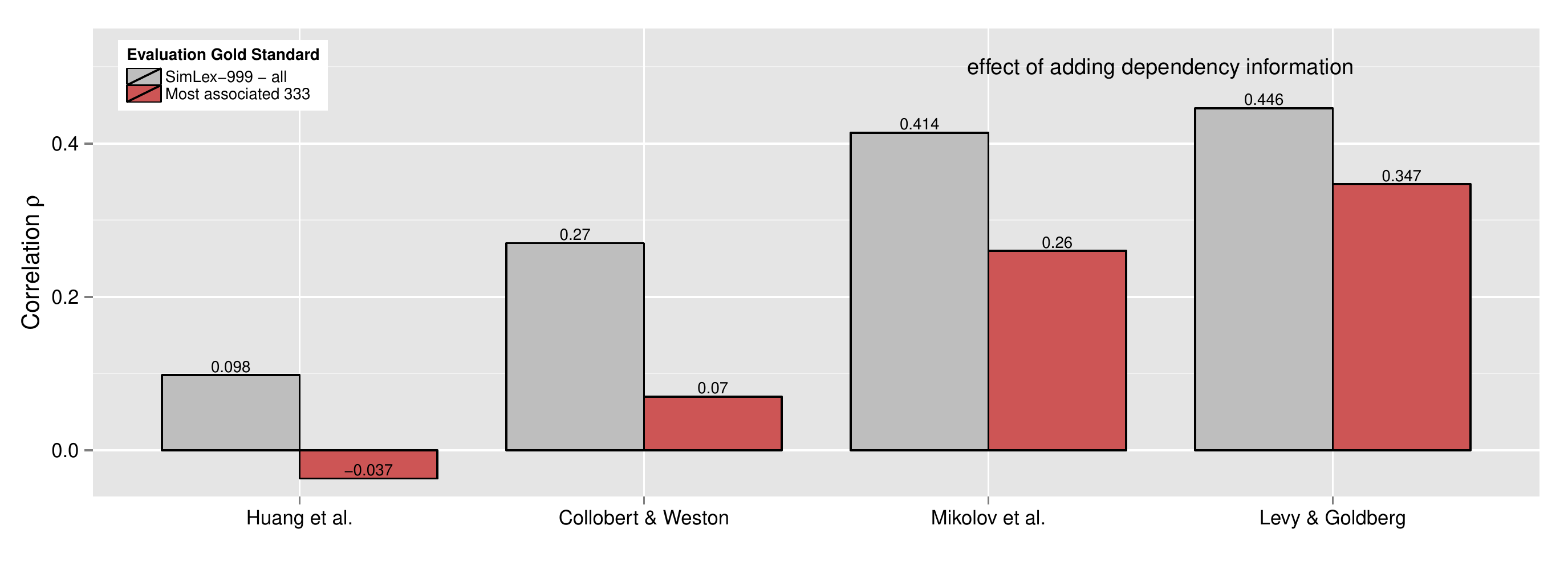}  \caption{The ability of NLMs to model the similarity of highly-associated concepts versus concepts in general. The two models on the right hand side also demonstrate the effect of training and NLM (the \protect\newcite{mikolov2013efficient} model) on running-text (\emph{Mikolov et al.}) vs. on dependency-based input (\emph{Levy \& Goldberg}).}\end{figure*}

\begin{figure*}[ht]  \includegraphics[width = \textwidth,height=4cm]{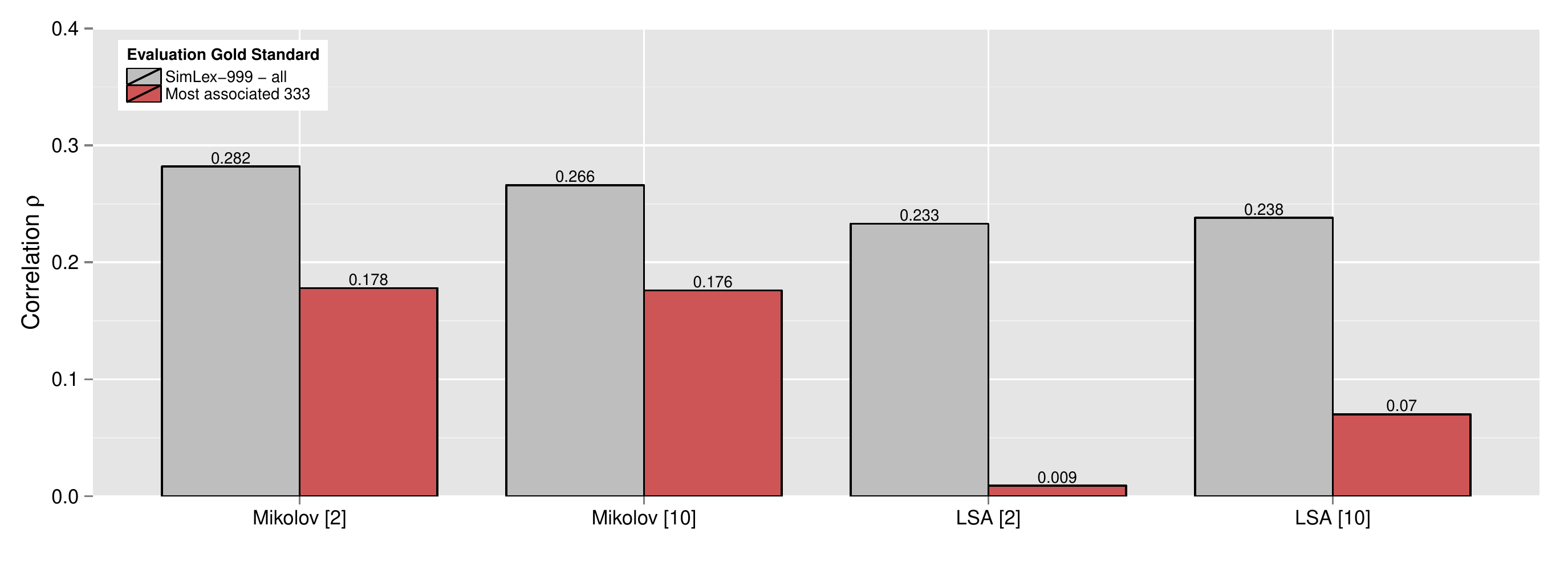}  \caption{The effect of different window sizes (indicated in square brackets [ ]) on NLM and LSA models . }\end{figure*}

The comparatively low performance of NLM, VSM and LSA models on SimLex-999 compared with MEN and WS-353 is consistent with our hypothesis that modelling similarity is more difficult than modelling association. Indeed, given that many strongly-associated but dissimilar pairs, such as [\emph{coffee, cup}], are likely to have high co-occurrence in the training data, and that all models infer connections between concepts from linguistic co-occurrence in some form or another, it seems plausible that models may overestimate the similarity of such pairs because they are `distracted' by association.

To test this hypothesis more precisely, we compared the performance of models on the whole of SimLex-999 versus its 333 most associated pairs (according to the USF free association scores). Importantly, pairs in this strongly-associated subset still span the full range of possible similarity scores (min similarity = 0.23 [\emph{shrink, grow}], max similarity = 9.80 [\emph{vanish, disappear}]).

As shown in Figure 8, all models performed worse when the evaluation was restricted to pairs of strongly-associated concepts, which was consistent with our hypothesis. The \newcite{collobert:2008} model was better than the \newcite{huang2012improving} model at estimating similarity in the face of high association. This not entirely surprising given the global-context objective in the latter model, which may have encouraged more association-based connections between concepts. The Mikolov et al. model, however, performed notably better than both other NLMs. Moreover, this superiority is proportionally greater when evaluating on the most associated pairs only (as indicated by the difference between the red and grey bars), suggesting that the improvement is driven at least in part by an increased ability to `distinguish' similarity from association.

\begin{figure*}[ht]  \includegraphics[width = \textwidth,height=4cm]{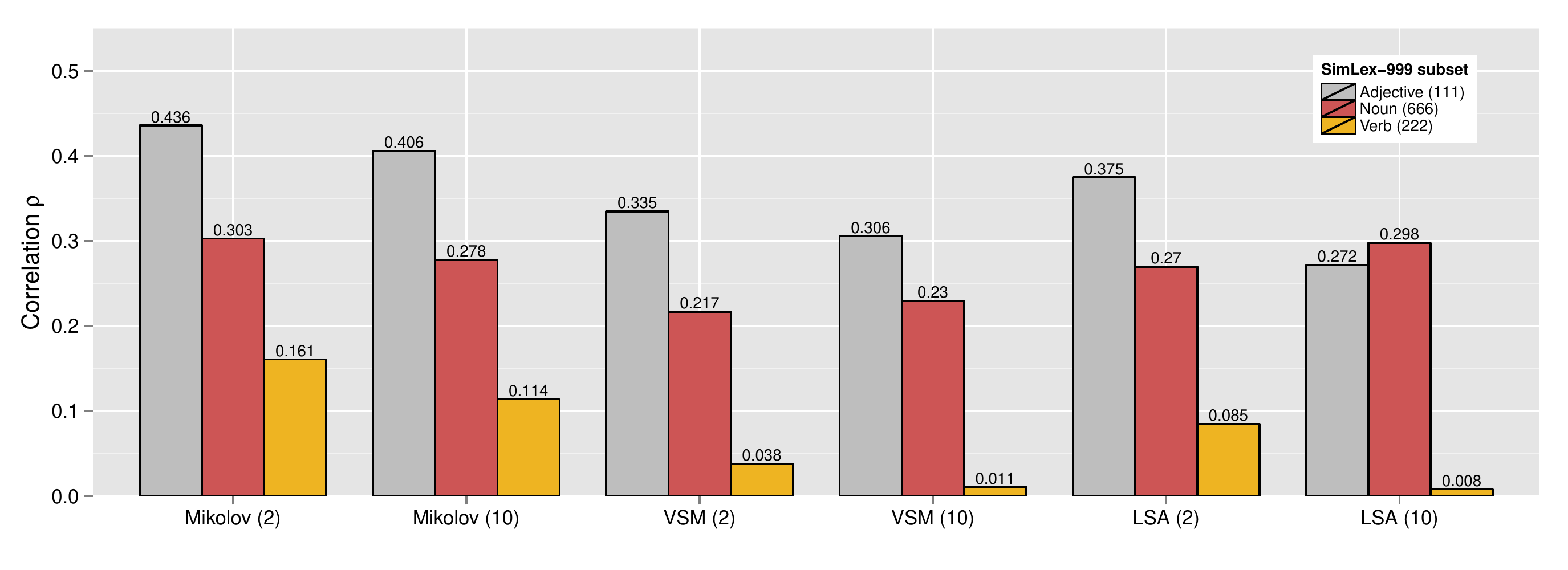}  \caption{Performance of models on POS-based subsets of SimLex-999. The window size for each model is indicated in parentheses.}\end{figure*}

\begin{figure*}[ht]  \includegraphics[width = \textwidth,height=4cm]{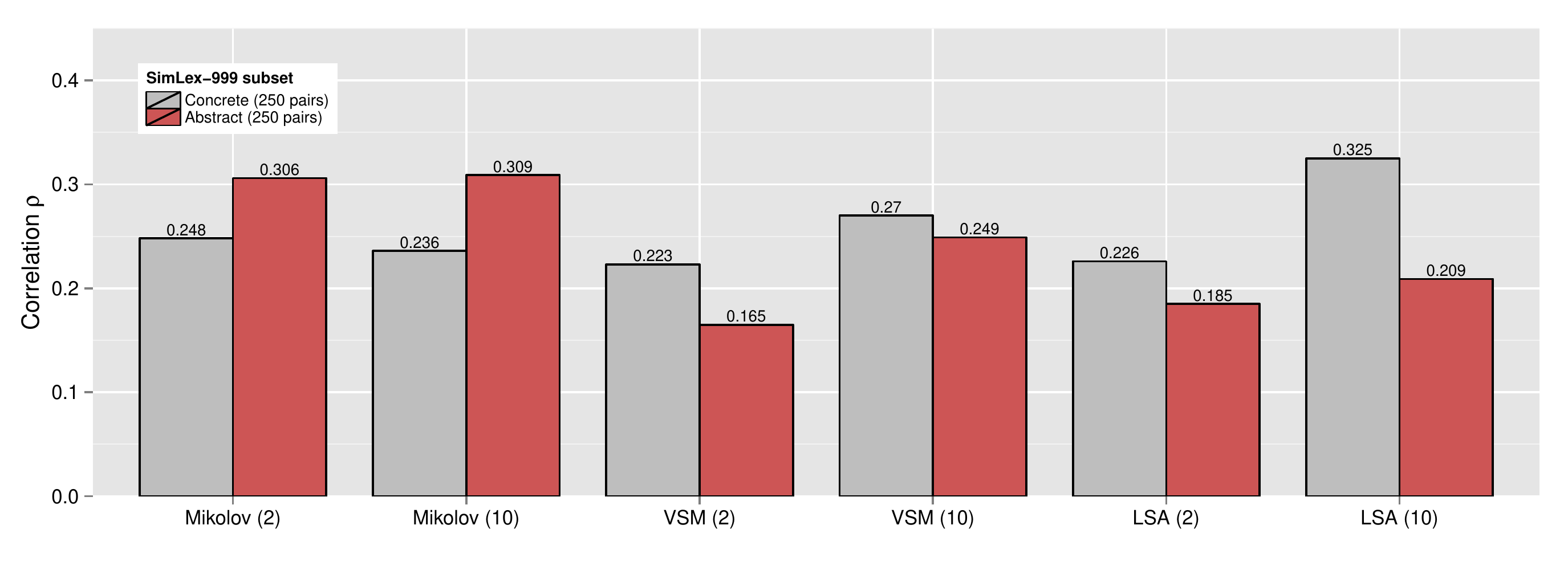}  \caption{Performance of models on concreteness-based subsets of SimLex-999. Window size is indicated in parentheses.}\end{figure*}

To better understand how the architecture of models captures information pertinent to similarity modelling, we performed two additional experiments using SimLex-999. These comparisons were also motivated by the hypotheses, made in previous studies and outlined in Section 2.1.2, that both dependency-informed input and smaller context windows encourage models to capture similarity rather than association.

We tested the first hypothesis using the embeddings of \newcite{levy2014dependency}, whose model extends the \newcite{mikolov2013efficient} model so that target-context training instances are extracted based on dependency-parsed rather than simple running text. As illustrated in Figure 8, the dependency-based embeddings outperform the original (running text) embeddings trained on the same corpus. Moreover, the comparatively large increase in the red bar compared to the grey bar suggests that an important part of the improvement of the dependency-based model derives from a greater ability to discern similarity from association.

Our comparisons provided less support for the second (window size) hypothesis. As shown in Figure 9, there is a negligible improvement in the performance of the \newcite{mikolov2013efficient} model when the window size is reduced from 10 to 2. However, for the LSA model we observed the converse. The LSA model with window size 10 slightly outperforms the LSA model with window 2, and this improvement is quite pronounced on the most associated pairs in SimLex-999.

\paragraph{\bf Learning concepts of different POS}

Given the theoretical likelihood of variation in model performance across POS categories noted in Section 2.2, we evaluated the \newcite{mikolov2013efficient}, VSM and LSA models on the subsets of SimLex-999 containing adjective, noun and verb concept pairs.

The analyses yield two notable conclusions, as shown in Figure 10. First, perhaps contrary to intuition, all models estimate the similarity of adjectives better than other concept categories. This aligns with the (also unexpected) observation that humans rate the similarity of adjectives more consistently and with more agreement than other parts of speech. Second, the effect of window size is also notable. A smaller context window is beneficial for learning both adjective and verb concepts, but this effect is not clearly observed for noun concepts.

We hypothesise that the smaller window size enables models to approximate the sort of inter-concept relationships that can otherwise be identified by dependency parsing. This is because the prior probability of a dependency relation existing between any two concepts within a small context window is discernibly higher than between two concepts within a larger window. This approximate dependency signal may be particularly important for learning adjective and verb concepts, which are sometimes referred to as \emph{relational concepts} \cite{markman1997similar} since they cannot typically be instantiated without other (normally nominal argument) concepts.

\paragraph{\bf Learning concrete and abstract concepts}

Given the strong interdependence between POS and conceptual concreteness (Figure 1), we aimed to explore whether the variation in model performance on different POS categories was in fact driven by an underlying effect of concreteness. To do so we compared performance of models on the most concrete and least concrete quartiles of the SimLex-999 dataset (Figure 11).

Interestingly, the performance of models on the most abstract and most concrete concepts suggests that the distinction characterized by concreteness is at least partially independent of POS. Specifically, while the Mikolov et al. model was the highest performer on all POS categories, its performance was worse than both the simple VSM and LSA models (of window size 10) on the most concrete concept pairs.

This finding supports the growing evidence for systematic differences in representation and/or similarity operations between abstract and concrete concepts \cite{hill2013concreteness}, and suggests that at least part these concreteness effects are independent of POS. In particular, it appears that models built from underlying vectors of co-occurrence counts, such as VSMs and LSA, are better equipped to capture the semantics of concrete entities, whereas the embeddings learned by NLMs can better capture abstract semantics.

\section{Conclusion}

Although the ultimate test of semantic models should be their utility in downstream applications, the research community can undoubtedly benefit from ways to evaluate the general quality of the representations learned by such models, prior to their integration in any particular system. We have presented SimLex-999, a gold standard resource for the evaluation of semantic representations containing similarity ratings of word pairs of different POS categories and concreteness levels.

The development of SimLex-999 was principally motivated by two factors. First, as we demonstrated, several existing gold standards measure the ability of models to capture association rather than similarity, and others do not adequately test their ability to discriminate similarity from association. This is despite the many potential applications for accurate similarity-focussed representation learning models. Analysis of the ratings of the 500 SimLex-999 annotators showed that subjects can consistently quantify similarity, as distinct from association, and apply it to various concept types, based on minimal intuitive instructions.

Second, as we showed, state-of-art the models trained solely on running-text corpora have now reached or surpassed the human agreement ceiling on WordSim-353 and MEN, the most popular existing gold standards, as well as on RG and WS-Sim. These evaluations may therefore have limited use in guiding or moderating future improvements to distributional semantic models. Nevertheless, there is clearly still room for improvement in terms of the use of distributional models in functional applications. We therefore consider the comparatively low performance of state-of-the-art models on SimLex-999 to be one of its principal strengths. There is clear room under the inter-rating ceiling to guide the development of the next generation of distributional models.

We conducted a brief exploration of how models might improve on this performance, and verified the hypotheses that models trained on dependency-based input capture similarity more effectively than those trained on running-text input. The evidence that smaller context windows are also beneficial for similarity models was mixed, however. Indeed, we showed that the optimal window size depends on both the general model architecture and the part-of-speech and concreteness of the target concepts.

Our analysis of these hypotheses illustrates how the design of SimLex-999 - covering a principled set of concept categories and including meta-information on concreteness and free-association strength - enables fine-grained analyses of the performance and parametrization of semantic models. However, these experiments only scratch the surface in terms of the possible analyses. We hope that researchers will adopt the resource as a robust means of answering a diverse range of questions pertinent to similarity modelling, distributional semantics and representation learning in general.

In particular, for models to learn high-quality representations for all linguistic concepts, we believe that future work must uncover ways to explicitly or implicitly infer `deeper', more general, conceptual properties such as intensionality, polarity, subjectivity or concreteness \cite{gershmanmetaphor}. However, while improving corpus-based models in this direction is certainly realistic, models that learn exclusively via the linguistic modality may never reach human-level performance on evaluations such as SimLex-999. This is because much conceptual knowledge, and particularly that which underlines similarity computations for concrete concepts, appears to be grounded in the perceptual modalities as much as in language \cite{barsalou2003grounding}.

Whatever the means by which the improvements are achieved, accurate concept-level representation is likely to constitute a necessary first step towards learning informative, language-neutral phrasal and sentential representations. Such representations would be hugely valuable for fundamental NLP applications such as language understanding tools and machine translation.

Distributional semantics aims to infer the meaning of words based on the \emph{company they keep} \cite{firth1957papers}. However, while words that occur together in text often have associated meanings, these meanings may be very similar or indeed very different. Thus, possibly excepting the population of Argentina, most people would agree that, strictly speaking, \emph{maradona} is not synonymous with \emph{football} (despite their high rating of 8.62 in WordSim-353). The challenge for the next generation of distributional models may therefore be to infer what is useful from the co-occurrence signal and to overlook what is not. Perhaps only then will models capture most, or even all, of what humans know when they know how to use a language.


\bibliographystyle{acl2012}
\bibliography{simlex}

\end{document}